\documentclass[11pt]{article}
\usepackage{graphicx}
\usepackage[margin=1.25in]{geometry}
\usepackage[usenames,dvipsnames]{color}
\usepackage{url}
\usepackage[colorlinks = true,
            linkcolor = blue,
            urlcolor  = blue,
            citecolor = blue,
            anchorcolor = blue]{hyperref}
\usepackage[sort,compress,numbers]{natbib}

\newcommand\pubnumber{Transcendental Preprint }
\newcommand\pubdate{\today}

\def\Title#1{\begin{center} {\LARGE #1 } \end{center}}
\def\Author#1{\begin{center}{ \sc #1} \end{center}}

\newcommand\pubblock{\rightline{\begin{tabular}{l} \pubnumber\\
         \pubdate \end{tabular}}}
\newenvironment{Abstract}{\begin{quotation} \begin{center}
                       ABSTRACT
     \end{center}\bigskip  }{\end{quotation}}

\def\beq{\begin{equation}}
      \def\eeq#1{\label{#1}\end{equation}}
\def\eeqn{\end{equation}}

\newenvironment{Eqnarray}%
{\arraycolsep 0.14em\begin{eqnarray}}{\end{eqnarray}}
\def\beqa{\begin{Eqnarray}}
      \def\eeqa#1{\label{#1}\end{Eqnarray}}
\def\eeqan{\end{Eqnarray}}

\def\lsim{\mathrel{\raise.3ex\hbox{$<$\kern-.75em\lower1ex\hbox{$\sim$}}}}
\def\gsim{\mathrel{\raise.3ex\hbox{$>$\kern-.75em\lower1ex\hbox{$\sim$}}}}

\def\del{\partial}
\def\Dslash{\not{\hbox{\kern-4pt $D$}}}
\def\dslash{\not{\hbox{\kern-2pt $\del$}}}
\def\pslash{\not{\hbox{\kern-2pt $p$}}}
\def\ETmiss{\not{\hbox{\kern-4pt $E$}}_T}

\def\Dlr{\mathrel{\raise1.5ex\hbox{$\leftrightarrow$\kern-1em\lower1.5ex\hbox{$D$}}}}

\def\MSB{{\bar{M \kern -2pt S}}}
\def\msb{{\bar{\scriptsize M \kern -1pt S}}}

\def\drb{{\bar{\scriptsize D \kern -1pt R}}}

\newcommand\snowmass{\begin{center}\rule[-0.2in]{\hsize}{0.01in}\\\rule{\hsize}{0.01in}\\
\vskip 0.1in Submitted to the  Proceedings of the US Community Study\\ 
on the Future of Particle Physics (Snowmass 2021)\\ 
\rule{\hsize}{0.01in}\\\rule[+0.2in]{\hsize}{0.01in} \end{center}}
\usepackage{acronym}
\acrodef{ML}{machine learning}
\acrodef{PF}{particle-flow}
\acrodef{GNN}{graph neural network}
\acrodef{LHC}{CERN Large Hadron Collider}
\acrodef{SONIC}{Services for Optimized Inference on Coprocessors}
\acrodef{HEP}{high energy physics}
\acrodef{HLT}{high-level trigger}
\acrodef{L1T}{level-1 trigger}
\acrodef{GAN}{generative adversarial network}
\acrodef{GCN}{graph convolutional network}
\acrodef{PID}{particle identification}
\acrodef{kNN}{k-nearest neighbors}
\acrodef{LSH}{locality sensitive hashing}
\acrodef{GPU}{graphics processing unit}
\acrodef{FPGA}{field-programmable gate array}
\acrodef{ECAL}{electromagnetic calorimeter}
\acrodef{HCAL}{hadron calorimeter}
\acrodef{PU}{pileup}
\acrodef{IPU}{intelligence processing unit}
\begin{document}

\pubblock

\Title{Physics Community Needs, Tools, and Resources for Machine Learning}

\bigskip

\Author{
	\textbf{Philip Harris, Erik Katsavounidis, William Patrick McCormack, Dylan Rankin}  \\
	Massachusetts Institute of Technology, Cambridge, MA 02139\\
	\textbf{Yongbin Feng, Abhijith Gandrakota, Christian Herwig, Burt Holzman, Kevin Pedro, Nhan Tran, Tingjun Yang, Jennifer Ngadiuba}  \\
	Fermi National Accelerator Laboratory, Batavia, IL 60510\\
	\textbf{Michael Coughlin}  \\
	University of Minnesota, Minneapolis, MN 55455\\
	\textbf{Scott Hauck, Shih-Chieh Hsu, Elham E Khoda}  \\
	University of Washington, Seattle, WA 98195\\
	\textbf{Deming Chen, Mark Neubauer}  \\
	University of Illinois at Urbana-Champaign, Urbana, IL 61801\\
	\textbf{Javier Duarte}  \\
	University of California San Diego, La Jolla, CA 92093\\
	\textbf{Georgia Karagiorgi}  \\
	Columbia University, New York, NY 10027 \\
    \textbf{Mia Liu}\\
    Purdue University, West Lafayette, IN 47907
}

\medskip

%\Address{ }

\medskip

\begin{Abstract}
	\noindent Machine learning (ML) is becoming an increasingly important component of cutting-edge physics research, but its computational requirements present significant challenges.
	In this white paper, we discuss the needs of the physics community regarding ML across latency and throughput regimes, the tools and resources that offer the possibility of addressing these needs, and how these can be best utilized and accessed in the coming years.
\end{Abstract}

\snowmass
\def\thefootnote{\fnsymbol{footnote}}
\setcounter{footnote}{0}

\pagebreak
\tableofcontents

\pagebreak

\section{Introduction \label{sec:intro}}

The field of machine learning (ML) has exploded in the last decades.
Many of the advances in the field have come hand in hand with advances in computing as algorithms have become increasingly complex, and ML is now in use across a wealth of research domains and industries.
In physics, specifically, ML has become a widely used tool for applications ranging from event classification to particle identification to energy regression.
New ML algorithms have improved both the performance and scope of these algorithms, but at the cost of algorithm complexity.
This evolution results in algorithms that can become computationally intensive or slow, potentially to an extreme degree.
As more complex algorithms are developed, and as the ML needs of physics research grow in the next decade, it will be vital for the physics community to be prepared with the computing tools and resources necessary to handle this growth.

Just as no single ML architecture is most appropriate for all problems, no single tool or resource for performing inference will address every physics use case effectively.
Indeed, as technologies and fields evolve so too will the methods most adapted to each use.
The goal of this white paper is not to lay out exactly how ML should be done for every physics application, but instead to present ideas that are capable of collectively addressing the needs of current and future physics experiments across frontiers.
These options come in many forms.
Hardware that can be faster and more efficient than traditional CPUs for inference is one possibility for reducing the overall computing load of ML.
Indeed in some cases specialized hardware is the only possible way to perform inference fast enough to be used effectively.
Other options for improved inference come in the form of software and computing paradigms.
These tools can improve the overall latency and throughput of inference and also reduce the computing complexity, significantly lowering the cost for users to develop optimized workflows.
Finally, there are multiple resource groups capable of enabling access to make use of these tools.
Cloud computing and in-house clusters are both options, as are high-performance computing (HPC) centers.
For environments that require microsecond-scale inference latencies, specialized resources for heavily leveraging FPGAs are necessary.
Taken together, and used effectively, these tools and resources can enable the novel ML that we believe will help drive the field of physics forward in the next decade.

The layout of this white paper is as follows.
Section~\ref{sec:needs} discusses the needs for ML in different frontiers, including colliders, neutrinos, and astrophysics.
Section~\ref{sec:tools} presents the different hardware and software tools that are available for accelerated ML and their advantages.
The necessary resources to properly utilize these tools for different use-cases are examined in Section~\ref{sec:resources}, and some potential exciting applications are presented in Section~\ref{sec:apps}.
A summary and outlook is provided in Section~\ref{sec:outlook}.

\section{Needs \label{sec:needs}}

\begin{figure}
    \centering
    \includegraphics[width=0.8\textwidth]{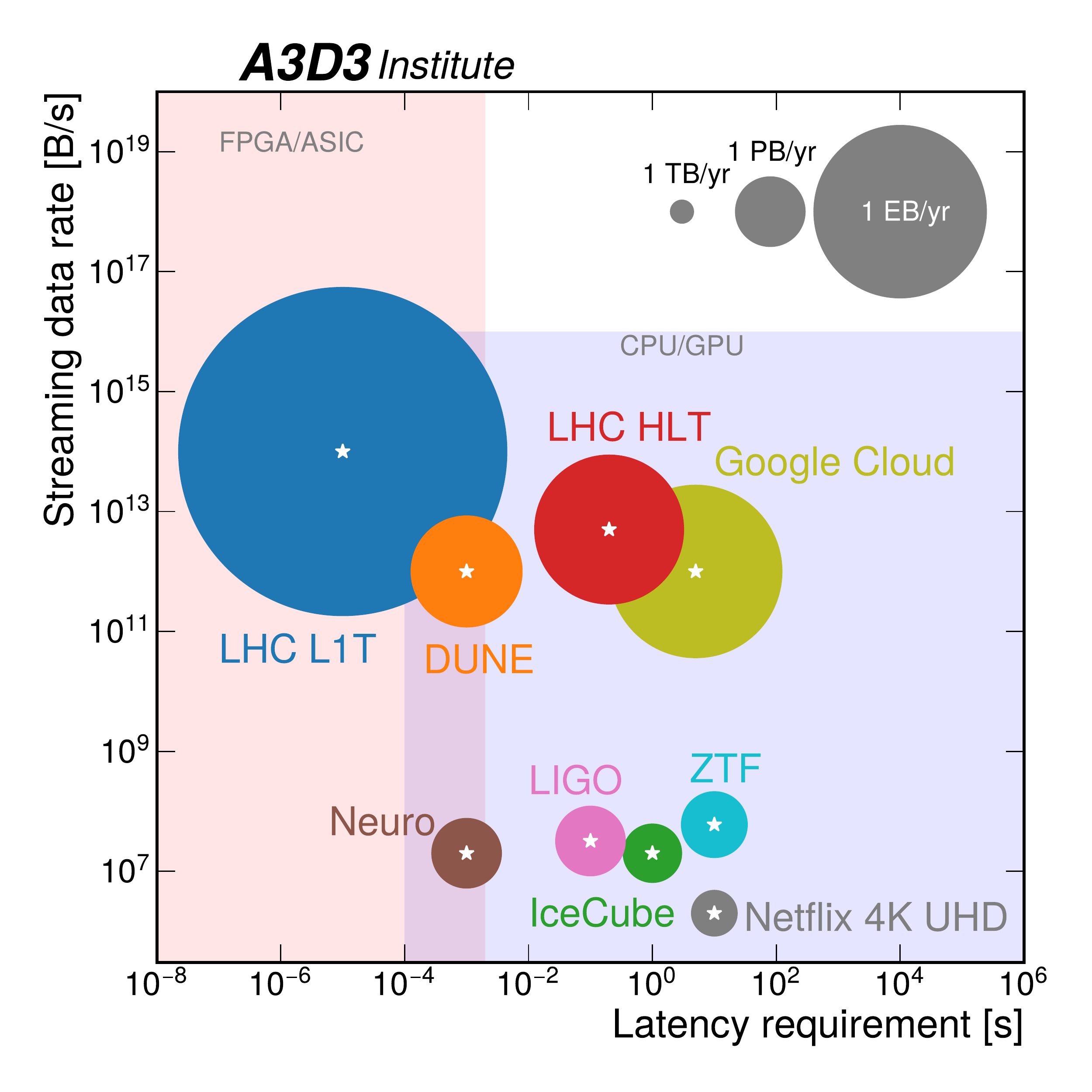}
    \caption{Plot of the streaming data rate in bytes per second and latency requirements in seconds for various experiments. Points of comparison from industry and other scientific fields are also included. The size of the bubbles represents the total per year data volume. Taken from Ref.~\cite{a3d3}.}
    \label{fig:a3d3}
\end{figure}

The ML needs of every experiment are inherently different.
These needs are dictated by many factors, such as the data formats, the system latencies, and the existing resources and workflows.
In order to frame the discussion of tools and resources we present the ML needs in three distinct fields (collider physics, neutrino physics, and astrophysics).
In each case ML has already begun to be widely adopted in certain contexts, and the needs are only expected to grow in the future.
A summary of the approximate data rates and latencies for experiments across these frontiers is shown in Fig.~\ref{fig:a3d3}.

\subsection{Collider Physics}

%Scale of algorithms + throughput needs. Intro both trigger and HLT/offline. \textbf{[1-2 paragraphs] Phil}
Data acquisition at collider experiments roughly fall into three separate tiers where reconstruction is applied at varying levels.
At each tier the latency and throughput demands of the reconstruction are quite different making the deployment of machine learning algorithms also quite different.
Moreover, the computing hardware between the different reconstruction tiers is also quite different. The combination of both of these demands make it such that the deployment of ML algorithms, and both the size and usage of ML algorithms varies greatly across reconstruction tier.

In the first tier of reconstruction at a collider experiment, data is often collected at an incredible rate.
At the Large Hadron Collider (LHC), collisions occur at a rate of 40 MHz.
With a single collision carrying several megabytes of raw information, overall data rates can approach a petabit per second. As a result of the large data rate and the harsh collision environment, the full event information can only be stored for roughly 10 microseconds~\cite{ATLASL1T,CMSP2L1T}.
As a result, an interconnected set of field-programmable gate arrays (FPGAs) are required to allow for the reconstruction of full events within the 10 microsecond window.
This stage of reconstruction is known as the Level-1 trigger (L1T).
As a result of these constraints, ML algorithms that are developed for this tier of reconstruction need to be small, and, depending on the reconstruction tier, to involve only local information.
Furthermore ML algorithms are required to run within the strict latency constraints, typically less than a microsecond.

By selecting, on average, the most interesting collision amongst one hundred, the first tier of reconstruction reduces the overall data rate by two orders of magnitude.
The second tier of reconstruction at a collider experiment, known as the high-level trigger (HLT), then takes this data and performs a much more thorough event reconstruction. Data selected for the HLT are shipped from the detector to a computing farm~\cite{Collaboration:2759072,CERN-LHCC-2017-020}.
As a result, the data are not stored in electronics subject to the harsh collider collision environment.
These conditions along with the reduced data rate allow for a loosening of the latency constraint from a fixed 10 microseconds to roughly 1 second or less in some cases.
Furthermore, the lower data rate allows for the use of more conventional CPUs and graphics processing units (GPUs) to perform the reconstruction.
The larger latency and access to conventional processing hardware allows for a large array of ML algorithms, provided they can be run within the latency constraints. Since these constraints are softer, algorithms can batch inputs together to process many collisions in parallel, allowing for the possibility of additional throughput gains when deploying these algorithms.

The final tier of reconstruction, typically referred to as offline reconstruction, takes in collisions selected by the HLT at a rate of a few kHz. At this tier, there is no latency requirement, and algorithms can take up to seconds per collision to run.
ML algorithms thus have no constraints.
Furthermore, recent developments in offline computing have allowed for the possibility of GPUs or other heterogeneous computing to be deployed for ML inference~\cite{Krupa:2020bwg}.

New generations of collider experiments are starting to pursue variations on these ML approaches. In particular, the LHCb experiment has largely eliminated the first tier of reconstruction and instead built a significantly larger HLT~\cite{LHCb:2018zdd}.
Additionally, future nuclear physics experiments are pursuing ``full-streaming'' readout, whereby they aim to read out all of the data in its raw form without a trigger, and then reconstruct the data offline over a substantially longer timescale~\cite{Ameli:2022gvw}.
These variations limit the need for FPGA-based or readout-based ML, at the cost of demanding more complicated, higher throughput ML computing downstream.

\subsection{Neutrino Physics}

%Scale of algorithms + throughput needs \textbf{[1-2 paragraphs] Georgia Karagiorgi}

In neutrino physics, ML and in particular applications of image recognition for neutrino identification have been growing \cite{Radovic:2018dip,Psihas:2020pby,Karagiorgi:2021ngt}, due to the increasing use of large, high-resolution tracking calorimeters as neutrino detectors.
Deep learning applications now span the full extent of data processing for neutrino experiments, including data acquisition \cite{Jwa:2019zlh,ArgoNeuT:2021xtd,Uboldi:2021jyj,ARIANNA:2021inb}, final data analysis (see, e.g.~Ref.~\cite{Aurisano:2016jvx,NOvA:2021smv,MicroBooNE:2021jwr}), and in particular data reconstruction (see, e.g.~Ref.~\cite{Baldi:2018qhe,MicroBooNE:2021ojx,KM3NeT:2020zod}).

A detector technology ideally suited for computer vision applications in neutrino physics is that of liquid argon time projection chambers (LArTPCs)---employed by the future DUNE \cite{DUNE:2020lwj}, current MicroBooNE \cite{MicroBooNE:2016pwy}, and upcoming SBN~\cite{MicroBooNE:2015bmn} experiments.
These detectors function as stereoscopic image streaming devices, offering the possibility of direct application of image recognition at as early as the data acquisition stage, for triggering purposes.
For the time being, however, applications of deep learning algorithms for these experiments have predominantly been focused on data reconstruction and final analysis tasks (see, e.g.~Ref.~\cite{DUNE:2020gpm,Liu:2020pzv,MicroBooNE:2016dpb,MicroBooNE:2020hho,MicroBooNE:2018kka,MicroBooNE:2020yze,Domine:2019zhm,SBND:2020eho,Drielsma:2021jdv,Adams:2020vlj}).
Beyond LArTPC experiments, other neutrino experiments such as NOvA \cite{Psihas:2017yuc}, MINERvA
\cite{MINERvA:2018smv}, Daya Bay \cite{Racah:2016gnm}, KamLAND-Zen \cite{KamLAND-Zen:2016pfg}, KM3NeT \cite{KM3NeT:2020zod}, and IceCube \cite{IceCube:2018gms,IceCube:2021dvc} are also making extensive use of ML for reconstruction and analysis purposes.

Event record sizes for neutrino detectors can be as large as 100 TB (for a supernova burst in the DUNE far detector), and typically of order of  $\leq1$~GB for the current generation of LArTPC experiments, challenging data reconstruction work flows.
With the use of increasingly larger and higher-resolution detectors, and the need for continuous readout for off-beam, rare-event physics searches, data readout and trigger challenges for neutrino experiments also begin to approach those of current collider experiments.
For example, the future DUNE \cite{DUNE:2020mra} will be generating raw data rates of several terabytes per second, and plans to be operated for at least a decade and with a 100\% live-time in order to be sensitive to supernova neutrinos and other rare and stochastic beam-unrelated signals.
New developments in ML applications for neutrino physics target GPU-accelerated ML inference as-a-service (see Sec.~\ref{sec:software}) for computing in neutrino experiments for data reconstruction purposes (see, e.g.~Ref.~\cite{Wang:2020fjr}), as well as GPU- or FPGA-based acceleration of ML algorithms such as 1D or 2D CNNs in real-time or online processing of raw LArTPC data at the data acquisition and trigger level \cite{Jwa:2019zlh,ArgoNeuT:2021xtd,Uboldi:2021jyj}.

% 20220321 Erik Katsavounidis: is the Neutrino Physics section going to cover high energy neutrinos from astrophysical sources (aka IceCube etc) and low energy neutrinos from CCSN?

\subsection{Astrophysics}

The growing number of astronomical instruments, including both ground- and space-based observatories, e.g. the Zwicky Transient Facility~\cite{Graham2019PASP}, the Submillimeter Array~\cite{HoMo2004}, the Very Large Array~\cite{Perley2011}, the Neutron star Interior Composition Explorer (NICER) ~\cite{Gendreau2016}, the Neil Gehrels Swift Observatory~\cite{Burrows2005,Roming2005}, amongst many others all across the electromagnetic spectrum, and starting recently also in gravitational waves ~\cite{GWTC3}, are challenging the existing data analysis paradigms.

So far, the usage of CPU and GPU clusters, combined with some level of human supervision, has been generally sufficient to perform the required source detection, classification, and parameter estimations. However, there are several considerations that call for a paradigm shift.
These include the exponential growth of data sets provided by the instruments, the interconnections between observations with all messengers (light, neutrinos, and gravitational waves), and the intrinsic timescales over which violent transient astrophysical sources with multi-messenger signatures develop, all of which make unattainable any human supervision/human-in-the-loop models, including requiring O(1s) latencies for processing data and disseminating results within the broader community for follow-up.

Some areas of transient astronomy have long been using machine learning techniques, e.g. Ref.~\cite{Bloom2012}.
However, newly established gravitational-wave astronomy with the ground-based km-scale interferometers like Advanced LIGO~ \cite{aLIGO}, Advanced Virgo~ \cite{adVirgo}, and KAGRA~ \cite{KAGRA} need a general infrastructure where ML models can be trained, deployed, and used for data analysis in a transparent way for the end-user, while factoring out the data acquisition and packaging protocols used in interferometry (see Ref.~\cite{Cuoco2020} for a summary of the various applications).
Altogether, it is the combination of providing accessibility and enabling ML algorithms with dedicated hardware that will yield the necessary automation, scalability, and latencies to meet the challenges and discovery potential of multi-messenger astronomy and astrophysics.

\section{Tools \label{sec:tools}}

The needs of physics experiments for fast and large-throughput ML inference will require a range of tools to address.
These tools are emerging from both industry and within physics.
We divide them into two categories: hardware and software.
The first category encompasses specialized hardware that is capable of improved performance with respect to traditional CPUs, as well as the software that exists for enabling the deployment of algorithms to run on these specialized processors.
The second category encompasses software capable of integrating ML inference into existing workflows and supporting at-scale computing of ML inference.
We also present lessons from the deployment of ML in industry.
We note that, although some tools for ML in industry may differ from those that are most useful in physics research, there is still much that can be learned.

\subsection{Hardware}

\begin{figure}[htpb]
    \centering
    \includegraphics[width=0.99\textwidth]{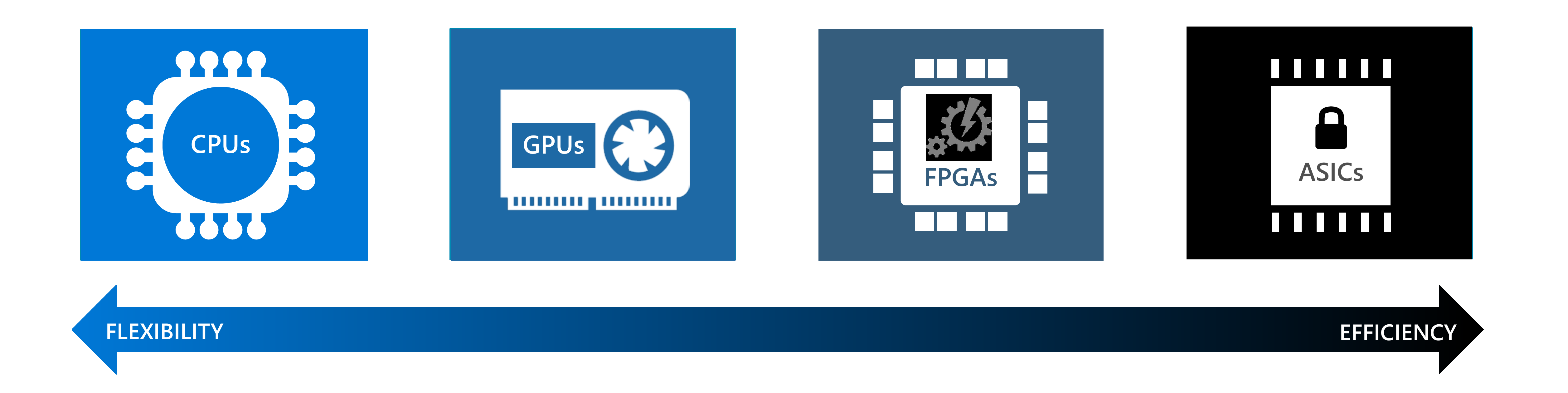}
    \caption{Silicon alternatives from most flexible (left) to most efficient (right), including CPUs, GPUs, FPGAs, and ASICs.
        TPUs (not pictured) are a relatively new development, which have flexibility similar to GPUs, but may have higher efficiency for ML inference. Figure adapted from Ref.~\cite{azuredocs}.}
    \label{fig:alternatives}
\end{figure}

While CPUs represent the core technology for computing, they are not particularly efficient for ML inference.
In order to enable low latency and high throughput inference alternative hardware will be a crucial addition to computing workflows.
A summary of the most common hardware is shown in Fig.~\ref{fig:alternatives}.
GPUs are the most well-known hardware capable of performing ML inference faster than CPUs.
Other alternative architectures like FPGAs, application-specific integrated circuits (ASICs), and new neuromorphic architectures (tensor processing units (TPUs)~\cite{TPU}, intelligence processing units (IPUS)~\cite{graphcore}, and more) have also become popular more recently due to their low power usage and speed.

\subsubsection{GPUs}

\begin{figure}[htpb]
    \centering
    \includegraphics[width=0.99\textwidth]{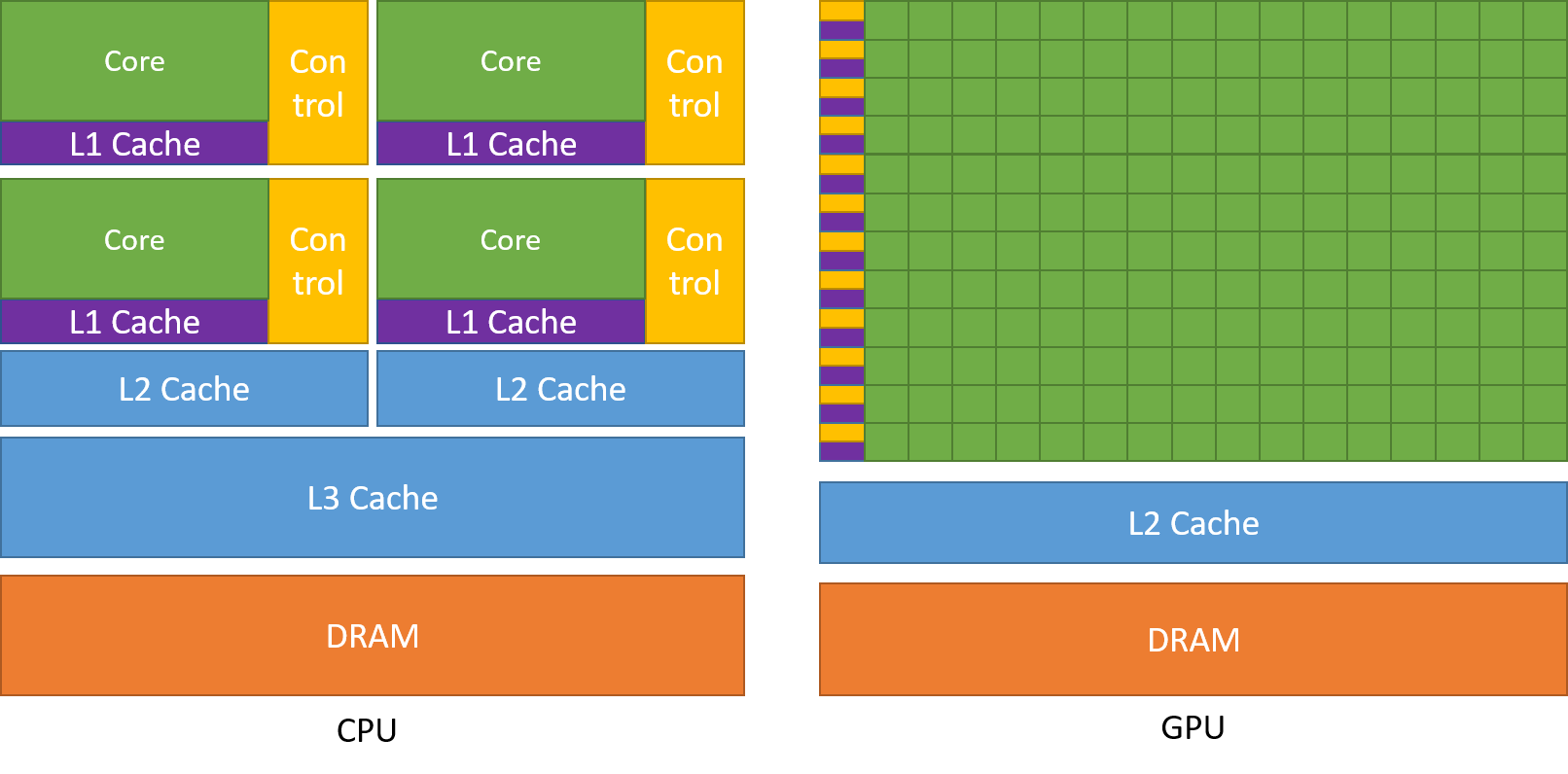}
    \caption{Architecture difference between CPUs and GPUs. Figure taken from Ref.~\cite{NvidiaCuda}.}
    \label{fig:cpu_vs_gpu}
\end{figure}

GPUs are specialized electronic circuits designed to rapidly manipulate and alter memory to accelerate the creation of images in a frame buffer intended for output to a display device. Figure~\ref{fig:cpu_vs_gpu} shows the architecture differences between CPUs and GPUs. Compared with CPUs, which are composed of a few cores with lots of cache memory good for serial processing, GPUs are composed of thousands of cores. This enables GPUs to perform \emph{parallel} operations on very large sets of data with much lower power consumption. While individual CPU cores are faster and more flexible, the large number of GPU cores and massive amount of parallelism make up for the single-core speed difference. GPUs are therefore well-suited for repetitive and highly parallel computing tasks.

Inference, as well as training, of ML models essentially consists of a large amount of complicated matrix operations. These operations are a very good fit for GPUs, especially when large batch sizes can be enabled. In the latest MLPerf inference benchmarks~\cite{gpuresults}, the throughputs on NVIDIA A100 GPUs are about $\mathcal{O}(100)$ times faster than Intel Xeon CPUs. The inference latency is typically around $\mathcal{O}(1)\,\mathrm{ms}$ to $\mathcal{O}(10)\,\mathrm{ms}$ per batch, well below the HLT latency requirement ($\mathcal{O}(100)\,\mathrm{ms}$). Accelerating both HLT and offline productions with GPUs is therefore a very promising direction.

Currently widely-used GPU programming languages include CUDA~\cite{cudalanguage}, OpenCL~\cite{DU2012391}, and OpenACC. Major ML frameworks are well-supported on GPUs, such as Tensorflow~\cite{tensorflow}, PyTorch~\cite{pytorch}, and ONNX~\cite{onnx}. Besides quantization and pruning (discussed in the following section), there are also well-established and supported tools to optimize and accelerate ML inference on GPUs, such as TensorRT~\cite{tensorrt}. These tools have the potential to dramatically increase the throughputs and reduce the cost.

\subsubsection{FPGAs}

\begin{figure}
    \centering
    \includegraphics[width=0.99\textwidth]{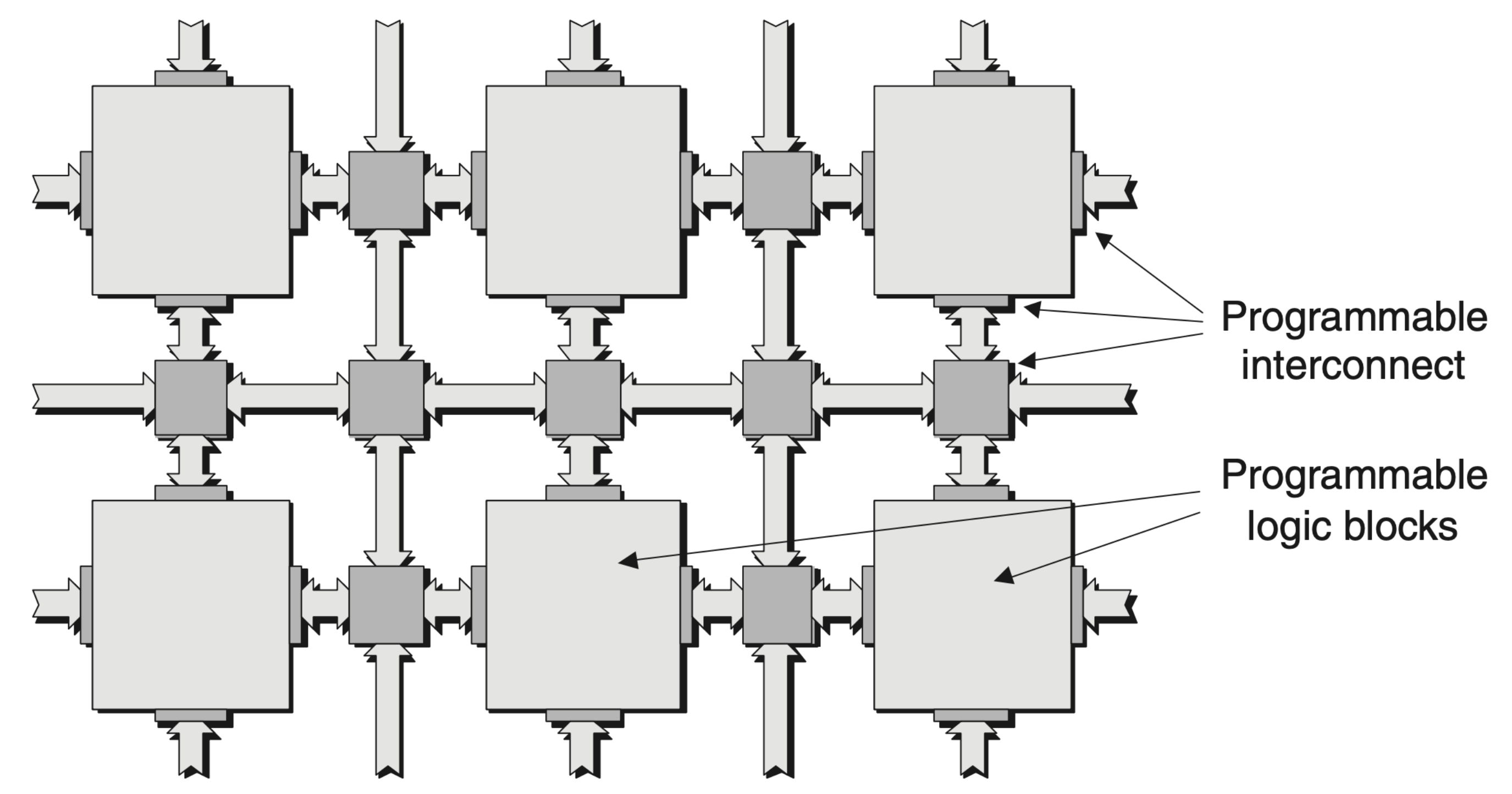}
    \caption{Simplified view of an FPGA architecture with programmable logic blocks and programmable interconnects.
        Figure adapted from Ref.~\cite{MAXFIELD200499}}
    \label{fig:programmable}
\end{figure}

Field-programmable gate arrays (FPGAs) are digital integrated circuits that contain configurable (i.e., \emph{programmable}) blocks of logic along with configurable interconnects between these blocks as shown in Fig.~\ref{fig:programmable}.
Algorithm designers can program such devices to perform an array of tasks.
Modern FPGAs also feature high bandwidth I/O connections and specialized components for multiplications (DSPs) or storing memory (block RAM).

Relative to CPUs, FPGAs can achieve much lower latencies for certain algorithms.
%For example, the inference latency of a benchmark deep learning algorithm like ResNet-50 can be accelerated to 1.9\,ms~\cite{DBLP:journals/corr/abs-2011-07317} compared to O(1\,s) or more on commercial CPUs without optimizations~\cite{MLPerfInference}.
In particular given the latency demands of Level-1 trigger systems at the LHC~\cite{CMSL1T,CMSP2L1T,ATLASL1T,ATLASP2L1T}, which are of order microseconds, an FPGA-based design is necessary.
Certain changes need to be made to the ML algorithms to enable efficient processing on an FPGA.
Namely, these are: \emph{quantization}~\cite{DBLP:journals/corr/HanMD15,DBLP:journals/corr/HanPTD15,DiGuglielmo:2020eqx,Coelho:2020zfu,Hawks:2021ruw}, or the use of reduced precision operations, e.g. 16-bit fixed-point precision instead of 32-bit floating point precision, and \emph{compression}~\cite{NIPS1989_250,DBLP:journals/corr/abs-1710-09282,9043731,review2020}, or the removal of unnecessary or redundant operations.
As an example, a reference benchmark autoencoder for anomaly detection runs on a CPU in 10\,ms, whereas a modified version of the same algorithm can be run on an FPGA in 9.6\,$\mu$s~\cite{Banbury:2021mlperf,tinyresults}.

Programming these devices requires highly specialized knowledge of register transfer-level (RTL) languages, like Verilog or VHDL, and vendor tools.
For this reason, custom source-to-source compilers, also known as transpilers, have been developed that can lower the barrier to entry for deploying ML algorithms on FPGAs by generating FPGA-ready firmware directly from trained ML models.
Many of these compilers leverage high-level synthesis (HLS)~\cite{surveyHLS,numan2020towards,hlsbook,ScaleHLS}, which is an alternative way of generating hardware modules from code written in high-level programming languages like \textsc{C}/\textsc{C++}.
Two such compilers are \texttt{hls4ml}~\cite{Duarte:2018ite} and FINN~\cite{blott2018finnr}, although many others exist in the literature~\cite{dnnweaver:micro16,venieris2016fccm,venieris2017fpga,venieris2017fpgaconvnet,venieris2017fpl,fpdnn,CloudDNN,DNNBuilder,PyLog}.
In particular, \texttt{hls4ml} has enjoyed wide usage within the particle physics community thanks to its flexibility and ease of use, especially for L1 trigger applications~\cite{CMSP2L1T}.
Current applications include binary and ternary neural networks~\cite{DiGuglielmo:2020eqx}, boosted decision trees~\cite{Summers:2020xiy}, quantization-aware training~\cite{Coelho:2020zfu}, convolutional neural networks~\cite{Aarrestad:2021zos}, graph neural networks~\cite{Iiyama:2020wap,Elabd:2021lgo}, and (variational) autoencoders~\cite{Govorkova:2021utb}.

Finally, FPGAs are not only applicable in the case of ultralow latency ML inference.
They also have use in datacenters for high-throughput inference tasks~\cite{Duarte:2019fta,Rankin:2020usv}, such as for processing of large experimental datasets.
This is discussed further in Sec.~\ref{sec:sonic}.

\subsubsection{ASICs}

For many particle physics experiments, the on-detector ``front end'' electronics consist of one or more ASICs working together to perform tasks such as data readout, calibration, and compression.
Often an ASIC is the only choice for these tasks, due to strict requirements on power consumption, radiation tolerance, and latency.
As detectors of the near future will generate data at unprecedented rates, these custom chips must be relied on to perform increasingly sophisticated processing in order for total recorded data volumes to be kept manageable, making ML algorithms a natural choice.

Here as in the case of FPGAs, specialized tools are required to port models trained with standard methods to a hardware design, specified in RTL.
However, recent capabilities of certain transpilation tools to specifically target ASIC architectures has enabled the use of similar toolkits such as \texttt{hls4ml} to specify completely custom circuits~\cite{catapulthls2020}.
In particular, this HLS flow enables quick prototyping so that designers can understand tradeoffs between model performance and the total chip area needed and estimated power consumption for their design.

Finally, the complete customizability of the ASIC provides for interesting options to allow for future flexibility of the ML model even once the final ASIC has been fabricated.
Ref.~\cite{DiGuglielmo:2021ide} illustrates an intermediate choice in which the CNN architecture is fixed, while weights may be reset via I2C registers, enabling reconfigurability to adapt to changing conditions over the lifetime of the experiment.
For applications requiring radiation tolerance, triplication of these weight parameters can ensure robustness to single-event effects~\cite{habinc2002functional,lyons1962use}.

\subsubsection{IPUs}

%In addition to more conventional GPUs, FPGAs, and ASICs, industry is also producing other specially-designed chips to accelerate ML-based applications. Intelligence Processing Units (IPUs) are one example of this class of processors. 

\begin{figure}[htpb]
    \centering
    \includegraphics[width=0.85\textwidth]{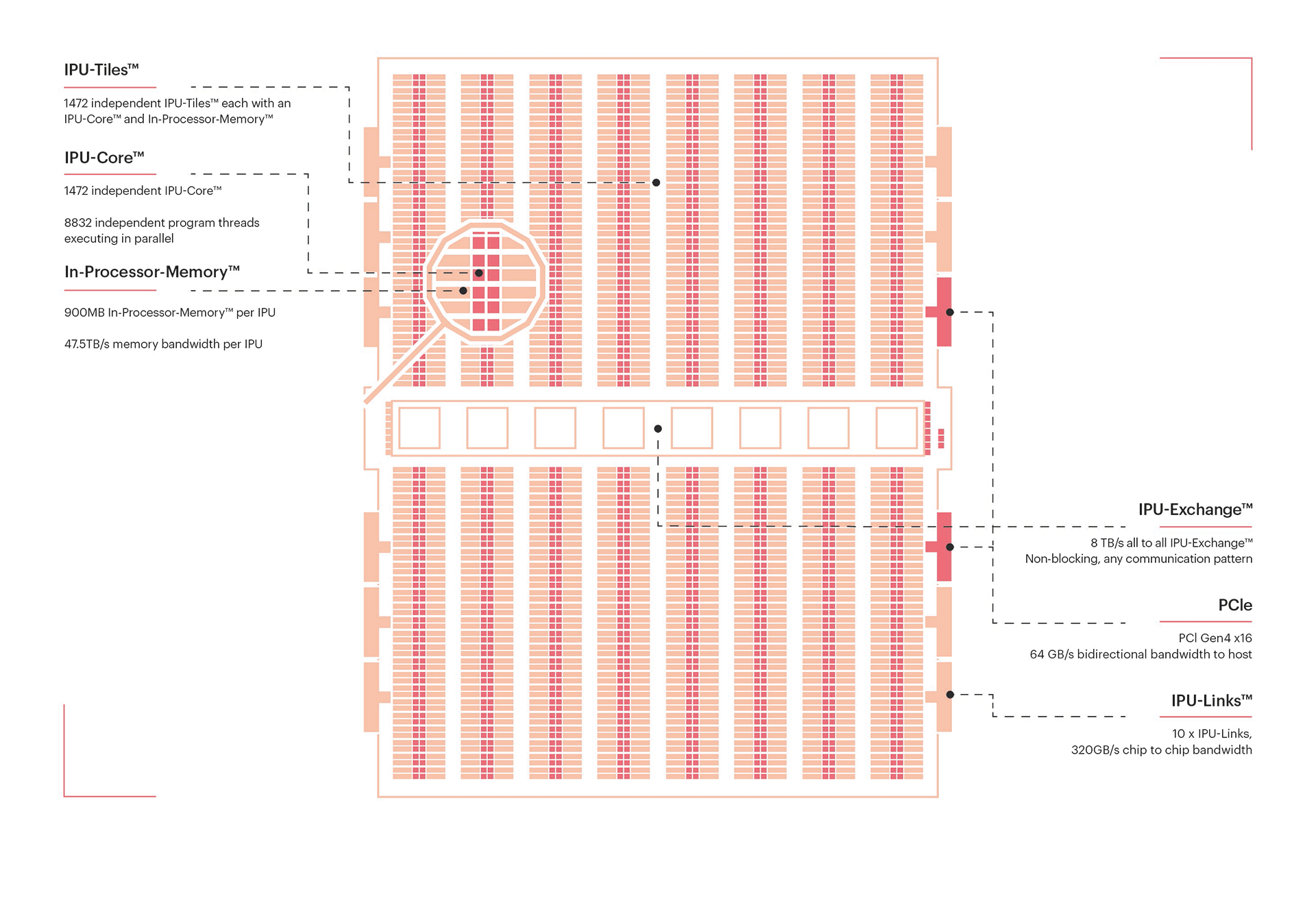}
    \caption{Simplified view of an IPU architecture. Figure taken from Ref.~\cite{graphcore}.}
    \label{fig:ipu}
\end{figure}

Intelligence Processing Units (IPUs) are massively parallel processors developed by GraphCore, with the design aim of efficient executions of fine-grained operations across a relatively large number of parallel threads~\cite{IPUPaper}. In contrast to GPUs, IPUs offer ``Multiple Instruction Multiple Data'' (MIMD) parallelism and adapt well to fine-grained, irregular computations that
exhibit irregular data access. Fig.~\ref{fig:ipu} provides a simplified view of an MK2 GC200 IPU architecture. Each IPU contains 1472 processing elements (``\emph{tiles}''); each tile consists of one computing core and around 600KB of local memory. IPU tiles communicate among themselves via \emph{links} and with CPU-based hosts via \emph{exchange}.

IPU programming is based on the Poplar SDK~\cite{Poplar}, co-designed with the IPU hardware. Currently the popular ML frameworks, such as Tensorflow~\cite{tensorflow2015-whitepaper}, PyTorch~\cite{pytorch}, and ONNX~\cite{onnx}, are well-supported with Poplar on IPUs. In the latest benchmark results released by GraphCore~\cite{IPUperf} on the training and inference of various popular ML models with different frameworks, IPUs outperform the state-of-art GPUs with higher throughputs and lower costs. Similar to GPUs, IPUs can potentially contribute to ML algorithm acceleration for both HLT and offline computing in HEP.

\subsubsection{Other specialized processors}

Other next-generation processor technologies are starting to emerge that could potentially dramatically improve the ability to perform ML computations. By restricting the processor technology to target these computations, optimized processors can be developed that explicitly target ML operations. One intriguing future possibility comes from optical based ML processors~\cite{Shen_2017,edgeoptical,Hamerly_2019}. Optical ML processors can operate at THz frequencies, leading to the possibility of running deep neural network inference within one nanosecond. Recent developments have allowed manufacturing these processors with conventional CMOS technology. Current optical-based technology is capable of running neural network interference with small networks ($\mathcal{O}(1,000)$ weights). Newer innovations could potentially allow for running inference for larger deep learning algorithms.

Tensor processing units (TPUs) are another type of optimized processor, designed by Google particularly for use with TensorFlow~\cite{TPU}.
Like many other specialized processors, they make heavy use of quantization techniques to reduce latency and resources.
TPUs are capable of processing hundreds of thousands of operations in a single clock cycle ~\cite{TPUblog}, and are optimized to perform operations like large matrix multiplications for training neural networks and image-based network inference, especially for very large networks. While TPUs are extremely performant at these tasks, these tasks do not represent a significant component of expected ML inference needs for physics.

\subsection{Software\label{sec:software}}

The architectures discussed above are capable of major performance improvements over standard CPU workflows, but achieving these performance improvements can be difficult.
Industry tools have been designed in many cases to provide fast inference for single processes and maintaining this performance across the potential range of architectures and system designs is difficult.
As a result of the popularity of GPUs, many industry tools are designed especially to make use of GPUs, but other tools exist from industry and in physics to make use of other architectures.
Integrating these tools and architectures into existing workflows can be non-trivial, and as a result methods to simplify this process are vital, particularly at production scale.
Paradigm shifts in computing such as the as-a-service model can enable more effective workflows not only for heterogeneous computing but also for CPU-only inference.

\subsubsection{Integration into existing workflows}

In a traditional computing model, a given CPU is responsible for processing the entirety of a given workflow, with each CPU capable of processing data in parallel to the others.
The integration of alternative architectures, or coprocessors, into this paradigm can take multiple different forms.
The simplest option conceptually is to connect a coprocessor to each CPU, and to use existing tools to offload part of the workflow to the coprocessor.
In the case that ML inference is a significant component of the workflow in terms of latency, the offloading of the inference to the coprocessor can provide a large reduction in the overall processing time.
While this is effective for smaller systems, its use in larger systems with many CPUs requires many coprocessors.
This can incur significant costs and also make it difficult to make full use of each coprocessor.
An alternative paradigm, called computing ``as-a-service'' (aaS), attempts to overcome these limitations by eliminating the need for direct connection between each CPU and coprocessor~\cite{EaaS,cloud_aas,service}.
In the aaS paradigm, coprocessors are connected to CPUs known as servers, and other CPUs known as clients communicate with the server CPUs over a network connection.
Clients directly perform the non-accelerated components of the workflow and send the inputs necessary to perform the accelerated components to the server.
This paradigm allows the management of the coprocessor to be separated from the clients, which can alleviate conflicts between the needs of ML inference and the needs of an existing system and workflow.
Additionally, the aaS paradigm simplifies the addition or removal of coprocessors from the workflow, which can be extremely beneficial as architectures improve or are phased out of service.

%\begin{itemize}
%    \item Comments on bare tf/pytorch here? \textbf{[1-2 paragraph] ???}
%    \item As-a-service \textbf{[1-2 paragraphs] Mia}
%    \item SONIC \textbf{[2-3 paragraphs] Kevin}
%    \item Triton (and other industry tools) \textbf{[1-2 paragraphs] Kevin}
%\end{itemize}

\subsubsection{Services for Optimized Network Inference on Coprocessors}\label{sec:sonic}

Services for Optimized Network Inference on Coprocessors (SONIC) is a software design pattern to integrate a client-server approach for inference as a service into experiment software frameworks (which are usually based on C++).
It offers useful abstractions to minimize dependence on specific features of the client interface provided by a given server technology.
SONIC has been implemented in the CMS software~\cite{Duarte:2019fta,Krupa:2020bwg,Rankin:2020usv} and in LArSoft for protoDUNE~\cite{Wang:2020fjr};
it is being explored by other experiments including ATLAS.
Asynchronous, non-blocking calls are the most efficient approach~\cite{Bocci:2020olh}, because the communication between the client and the server proceeds in parallel with other work continuing on the local CPU, therefore hiding the impact of transmission latency.
However, synchronous, blocking calls may still provide significant speedups if a workflow is dominated by ML algorithm inference that can be offloaded to faster coprocessors.
Here, we provide a brief overview of the advantages of SONIC.
More details can be found in the aforementioned references, which also provide performance results showing speedups by more than an order of magnitude for a variety of ML algorithms.

\begin{itemize}
    \item \textbf{Flexibility}: Allowing multiple clients to connect to multiple coprocessors enables many arrangements to ensure optimal usage of all devices.
    \item \textbf{Cost-effectiveness}: Related to flexibility, using coprocessors optimally reduces the number of coprocessors that must be purchased to support algorithm inference.
    \item \textbf{Symbiosis}: SONIC facilitates the use of existing industry tools and developments (see Section~\ref{sec:industry_tools}), rather than requiring HEP software developers to reimplement common tasks such as ML algorithm inference repeatedly for different ML frameworks, coprocessors, etc.
    \item \textbf{Simplicity}: SONIC modules only implement conversions of input and output data, which reduces the amount of code necessary to develop and maintain in order to perform ML algorithm inference.
    \item \textbf{Containerization}: The client-server model keeps the ML frameworks separate from the experiment software framework, which eliminates the significant workload needed to integrate two software systems that each have their own complicated dependencies.
    \item \textbf{Portability}: SONIC enables experiment software workflows to swap between CPUs, GPUs, FPGAs, and other coprocessors such as IPUs without any code changes. In some cases, even the choice of ML framework can be changed with no other modifications.
\end{itemize}

It should be understood that the flexibility of SONIC is solely an advantage, in that it provides options, but does not require them to be used.
In particular, SONIC facilitates the use of co-located or remote (e.g. cloud) coprocessor resources that may not be directly connected to worker CPUs.
However, many of the other advantages of SONIC can still be realized even if it is just used as an abstraction to interact with directly-connected coprocessors.
SONIC can also offload non-ML algorithms, but this may lack the advantage of automatic portability that comes with ML.
While the use of separate servers and/or separate inference processes does add some complexity to experiment workflows, this is of a similar magnitude as, for example, conditions databases or remote data access protocols that are already widely used.

\subsubsection{Industry tools}\label{sec:industry_tools}

Existing implementations of SONIC~\cite{Krupa:2020bwg,Wang:2020fjr} focus on the open-source Triton inference server from Nvidia~\cite{Triton}.
It offers a number of useful features, including:
\begin{itemize}
    \item Support for all modern ML frameworks
    \item Support for non-ML algorithms and even non-Nvidia coprocessors via custom backends loaded by the server
    \item Standardized protocols, http or gRPC~\cite{gRPC}, to communicate between the client and the server
    \item Load balancing for multi-GPU servers
    \item Dynamic batching: combining multiple requests for the same algorithm to be processed more efficiently in GPU memory
    \item Usage of shared memory for faster communication with directly-connected processors (CPU or GPU)
    \item Input/output compression to conserve bandwidth
    \item Tools to optimize algorithm deployment on the server
\end{itemize}
The extensible nature of the server, with its standardized protocols and custom backends, has already been used to deploy servers that conduct inference on FPGAs or IPUs rather than Nvidia GPUs.
This enables the automatic portability that is a key advantage of the SONIC approach.
Previous implementations of SONIC have used the Microsoft Brainwave API~\cite{caulfield2016a} as well as wrapping the TensorFlow C++ API.
In the future, other client-server technologies such as the interprocess communication (IPC) provided by Apache Arrow~\cite{arrow} could be considered.

The use of industry client-server technologies enables, though it does not require, the use of commercial cloud resources and associated software tools.
All major providers---Google Cloud, Amazon Web Services, and Microsoft Azure---have been used successfully to host coprocessor servers that provide GPUs or FPGAs.
The option to use cloud computing is also useful to test new devices such as IPUs~\cite{graphcloud}.
Kubernetes~\cite{k8s} is an extremely powerful and versatile tool to orchestrate these cloud servers, and it is therefore worth investigating seriously for HEP-specific grid computing centers as well.

\subsection{Lessons from industry}

Significant effort in industry has been dedicated to developing software frameworks, computing and networking systems, and hardware accelerators to meet the broad range of demands for ML solutions in various fields. These range from smart Internet of Things (IoT) domains (e.g., smart city, smart manufacturing, wearable healthcare devices, and security surveillance) to computer vision (e.g., autonomous driving, unmanned aerial vehicles, and augmented/virtual reality) to data analytics applications (e.g., those encountered in particle physics for scientific discovery). %Industry has created a healthy, vibrant and evolving eco-system where ML algorithms can be efficiently developed in easy-to-use open-source frameworks (e.g., PyTorch, TensorFlow, and Caffe) and then smoothly mapped to CPUs, GPUs, or TPUs for acceleration targeting different use cases. 
Many industry tools focus on ease of use, and as such development flows have been streamlined and been made highly efficient.
The importance of these easy-to-use frameworks and tools such as PyTorch~\cite{pytorch}, TensorFlow~\cite{tensorflow2015-whitepaper,tensorflow}, and ONNX~\cite{onnx} remaining open source is of vital importance to the physics community.
Tools that are not open source can be difficult to use due to a lack of ability to adapt to the specifics of physics workflows.

For low-latency solutions that may require FPGAs as hardware accelerators, the commercial HLS tools from major FPGA vendors, such as AMD/Xilinx and Intel/Altera, are of critical importance. The broad usage of FPGAs (and ASICs) for ML in physics is highly dependent on these HLS tools. These vendors also offer ML-specific design flows. For example, AMD/Xilinx offers the Vitis AI development environment for AI inference on AMD/Xilinx hardware platforms, that contains optimized IP, tools, libraries, models, and example designs. It also offers AI Model Zoo, which provides optimized and retrainable AI models for fast deployment and high-performance accelerations.
It is useful to note that the low-latency regime for industry is typically focused on latencies of approximately 1 ms.
While these latencies are appropriate for many physics applications, there also exists cases with latency needs in the sub-microsecond regime, far below the target of most industry tools.

Industry has seen much success through efforts coordinated between users, application developers, hardware vendors, and computing and networking solution providers to help create a highly functional ecosystem. This is essential for the fast growth of ML-driven businesses. The scientific community can learn from this and carry out coordinated efforts among researchers and practitioners in physics, computer science, engineering, and hardware systems to achieve our ultimate research goals. We should also continue to develop dedicated and customized HLS solutions, compilers, and hardware systems in order to achieve high computation efficiency and low latency for the ML algorithms specific to our scientific domains.

\section{Resources \label{sec:resources}}

While some physics workflows have already begun to incorporate coprocessors, many are still limited only to CPUs.
The variety of resources for accessing coprocessors will be important as ML inference and computing needs expand.
While some experiments are able to make use of more generic options, many applications will require special resources.

\subsection{On- and off-detector real-time electronics}

% Discussion of resources needed for triggers, both full design and prototyping \textbf{[1-2 paragraph] Nhan}

Machine learning integrated into the detector real-time electronics provides powerful data reduction capabilities, particularly for very high data rate or low latency applications at the LHC, DUNE, particle accelerators, and many more~\cite{deiana2021applications}. We are considering here applications ranging from on-detector data concentrator or aggregation in ASIC or FPGAs to off-detector real-time trigger or filtering based on feature extraction of physics observables. To consider such a wide range of custom requirements for each specific application---with different bandwidths, latencies, interfaces, hardware platforms---makes developing common resources challenging. Therefore we broadly consider two types of resources needed for the development of custom electronics ML solutions: hardware platforms and electronic design automation (EDA) tools.

Hardware platforms can vary from custom ASIC chips to custom FPGA systems to off-the-shell electronics boards. Custom hardware solutions are tailored to specific experiments, particularly in the case of ASICs, but there are some FPGA readout systems than can serve multiple experiments such as OTSDAQ/CAPTAN~\cite{biery2018fermilab,4775101}, RCE~\cite{7431254}, and FELIX~\cite{8700221}. Beyond that, flexible off-the-shelf solutions for use-cases with less stringent requirements are becoming increasingly popular. There are a wide range of examples that make compiling a complete list challenging. They can range from commercially available system-on-chip (SoC) or system-on-module (SoM) partial solutions that can be integrated into a larger system---for example, an Arria10 SoM~\cite{reflexces} being explored for accelerator controls---to development kits to complete solutions like the Alveo~\cite{alveo}. The Alveo is even available in the cloud~\cite{f1instance} and can be used for prototyping.

Resources for electronics synthesis and integration can be extremely expensive, including maintaining licenses and also support more generally. In the previous section, we discussed tools for ML translation to hardware description language; however, integrating those algorithms into FPGA or ASIC systems requires licenses for expensive EDA tools.  FPGA tools such as Vivado~\cite{vivado} and Quartus~\cite{quartus} are not relatively expensive compared to ASIC tools, but are much more widely used, and university and laboratories need to reserve resources to maintain licensing of those tools. For ASIC development, the cost can be an order of magnitude or more greater for EDA tools (which can also synthesize designs for FPGAs) from vendors like Cadence~\cite{cadence}, Synopsys~\cite{synopsys} and Siemens~\cite{mentor}. Two potential paths that should be supported to reduce resources include: (a) joint agreements across laboratories and universities and the vendors to reduce overall costs and (b) exploration of open-source solutions.  For (b) in particular, the trend towards open-source hardware both for FPGAs and ASICs continues to grow with the development of tools like Symbiflow~\cite{symbiflow}, Yosys~\cite{yosys}, and SkyWater SKY130 PDK~\cite{skywater}.

\subsection{Cloud providers \label{sec:cloud}}

Perhaps the simplest way to acquire large-scale computing resources on a short timescale is the utilization of cloud resources, such as Google Cloud Platform (GCP)~\cite{GCP} or Amazon Web Services (AWS)~\cite{AWS}.
By coordinating with the provider, it is straightforward to acquire $\mathcal{O}(10,000)$ CPU threads and $\mathcal{O}(100)$ GPUs for use at a given time.  There is also flexibility to choose from a variety of CPU platforms and GPU types~\cite{GCP_CPU, GCP_GPU, AWS_EC2}, change the platform at will, and create custom images to deploy in all instances.

To manage these large-scale resources, one can use pre-existing paradigms, such as \texttt{HEPCloud}~\cite{hepcloud}, which emulates the HTCondor distributed computing system, dynamically allocating and removing worker nodes in the cloud as needed.
It is also possible to build a custom \texttt{SLURM} workload manager in the cloud. The HEPCloud approach allows for greater ease of access, as relatively little user-side expertise is needed, while the \texttt{SLURM} approach allows for somewhat more control over resource configuration and the use of GPUs in worker nodes, though a greater deal of user-side effort and knowledge is required to correctly configure cluster networks and images.

%While the cloud provides for quick large-scale access to computing resources, it is likely best employed for relatively short-term development and testing purposes; for example, in the testing of a new architecture or device, or in scale-up tests of a particular computing scheme.
%Cloud computing costs can be prohibitive for long term running, with costs for CPU-only nodes potentially in the range of $\mathcal{O}(\$ 1- \$ 10)$ per day per thread, depending on the memory, disk, and platform specifics, and GPU-enabled nodes costing $\mathcal{O}(\$ 1,000)$ per month for continual operation. 
While the cloud provides for quick large-scale access to computing resources, costs can be prohibitive for long term running. Rates for CPU-only nodes are in the range of $\mathcal{O}(\$ 1- \$ 10)$ per day per thread, depending on the memory, disk, and platform specifics, and GPU-enabled nodes cost $\mathcal{O}(\$ 1,000)$ per month for continual operation. 
The cost of CPU-enabled nodes can be reduced by a factor of about 5 by using pre-emptible or ephemeral nodes (which can be ``taken'' by other users with higher priority), though the feasibility of this option depends on the use case.

In spite of this high cost, there are certain cases that are very well-suited to the cloud.
Short-term development and testing can make effective use of the pool of heterogeneous resources, without requiring full purchases of a resource before its potential is understood.
Several proof-of-principle studies have demonstrated the possibility of deploying heterogeneous computing platforms, such as GPU and FPGA coprocessors, in HEP software with an innovative ``as-a-service'' approach~\cite{Krupa:2020bwg, Rankin:2020usv}.
The SONIC software framework was developed during these studies, which were made possible through large-scale heterogeneous cloud resources, and the SONIC-based GPU- and FPGA-as-a-service toolkits have been released as open source code.
Scale-up tests of computing scenarios are another scenario for which the cloud is an appealing option.
The cloud can be used in these cases to make heavy use of a huge amount of resources for a short period of time.
For example, an initial test using a small number of simulated ProtoDUNE events showed a viable, cost-effective way to use the SONIC framework to solve the computing challenges facing the neutrino experiments~\cite{Wang:2020fjr}. 
In 2021, to take advantage of an improved machine learning model, the entire 7 million ProtoDUNE beam data events were reprocessed using this framework. The acceleration provided by GPUs accessed through Google Cloud reduced the total processing time by half compared to the CPU-only approach.
Similar tests have also shown the possibly of large speed-ups for gravitational-wave data processing by using cloud resources~\cite{Gunny:2021gne}.

In the future, it is likely that cloud computing costs will decrease. Cloud computing providers are able---in principle---to procure computing resources at a fraction of our cost due to the very large scale of their procurement processes.  They also can benefit from economies of scale in operation and development.  Additionally, they rent computing power in a free market (sized at more than \$300 billion), competing with not only on-premises resources but other cloud computing companies.

\subsection{High Performance Computing (HPC) Centers}

There are a number of large HPC centers supported by the National Science Foundation and the Department of Energy. While many of these facilities are optimized for traditional high-performance computing jobs (requiring low-latency communication between nodes), they are also capable of executing high-throughput computing jobs (requiring little or no intranode communication). Work is executed using a fair-share scheduler. There are several challenges with federating these resources across multiple sites. The network connectivity between the compute nodes and the outside world may be underprovisioned (or in some cases, non-existent). Some systems are secured with multi-factor authentication and many do not have API endpoints for workload submission available off-site. Many centers use an annual proposal/review/award allocation model, which is not a good fit for HEP experiments that can process data over the course of decades.

Some HPC facilities deploy significant numbers of GPUs (e.g. Summit \@ Oak Ridge National Laboratory, 27k NVIDIA V100 GPUs; Perlmutter \@ NERSC, 6,000 NVIDIA A100 GPUs). Training of machine learning models, which can make efficient use of GPU to CPU ratios of 2:1 (or significantly more, depending on the model), can be executed on the machines using standard fair-share scheduling. However, processing data with the inference-as-a-service model---which may only need GPU to CPU ratios of 1:300 or less~\cite{Krupa:2020bwg}---requires different co-scheduling of GPU and CPU resources. Addressing this difference may require direct coordination with HPC staff and changes to the facility scheduling model.

%Some facilities: Frontera and Stampede2 (Texas Advanced Computing Center), Bridges (Pittsburgh Supercomputing Center (PSC), Cheyenne, at the NCAR-Wyoming Supercomputing Center, Expanse, at the San Diego Supercomputer Center (SDSC), Jetstream, at the Indiana University Pervasive Technology Institute (PTI),  NERSC (LBL), ALCF (ANL), OLCF (ORNL)

\section{Applications \label{sec:apps}}

Although by no means a comprehensive summary, we offer a select few applications of ML that can be enabled through the effective use of the tools and resources described above.
As in Section~\ref{sec:needs}, we divide the applications by frontier.

\subsection{Colliders}

%Highlight L1 trigger possibilities. Clustering? Tracking? \textbf{[1 paragraph] Shih-Chieh}

%Machine learning technologies offer promising solutions and enhanced capabilities in collider physics both in real-time and offline, thanks to their capacity for extracting the most relevant information from high-dimensional data and to their highly parallelizable implementation on suitable hardware.
%With the scheduled upgrades in 2025, the data rate at the \ac{LHC} will approach a petabit per second, far exceeding the storage and analysis capacity of any device in the world, thereby demanding the design and use of real-time methods throughout the data stream~\cite{Alves:2017she}.
%At the LHC, data acquisition and reduction is done in three tiers: the \ac{L1T}, the \ac{HLT}, and offline computing, spanning processing latencies of less than 10\,$\mu$s, hundreds of milliseconds, and seconds, respectively. Each successive tier selects $\sim$\,1/100 of the most interesting collisions for further analysis. 
%Real-time and accelerated ML are expected to provide a great solution to efficiently handle the computationally-intensive processing of upgraded LHC data.

The stringent latency and resource constraints imposed by the L1 trigger at the LHC make it an ideal target for applying innovative ML methods to embedded systems.
%There have been several examples of the application of machine learning models to physics tasks at the LHC, together with novel methods for their efficient deployment in both the real-time and offline data processing stages.
%Machine learning models designed for the L1 trigger 
There have been several examples of the application  of machine learning models based on high-level synthesis tools~\cite{Duarte:2018ite,Aarrestad:2021zos,Loncar:2020hqp,Summers:2020xiy,Hong:2021snb, Rao:2729154} from FPGA vendors for tasks such as the reconstruction and calibration of final objects or lower-level inputs like trajectories~\cite{Elabd:2021lgo}, vertices, calorimeter clusters~\cite{CMSP2L1T}, and identification of long-lived particles~\cite{Alimena:2020web}. Alternative approaches are being considered based directly on hardware description languages, such as VHDL~\cite{Nottbeck:2020nke}, for example, the real-time signal processing of the ATLAS Liquid Argon calorimeter~\cite{Laatu:2789991,Chiedde:2021bmq}.
Anomaly detection techniques such as autoencoders are being explored to efficiently suppress the SM background contribution without imposing stringent kinematic constraints~\cite{Govorkova:2021utb,Mikuni:2021nwn,Jawahar:2021vyu,Butter:2022lkf}.
Deploying such a triggering mechanism at L1 trigger on FPGAs using the \texttt{hls4ml} tool can significantly enhance the sensitivity to new physics and rare SM processes.
Modern architectures such as graph neural networks (GNNs) are being explored for the reconstruction of particle trajectories, showers in the calorimeter as well as of the final individual particles in the event.
Key GNN applications can only be realized at scale with optimized hardware inference for GNNs, which is a innovative topic within the field of ML algorithm-hardware co-design.

Accelerated ML can be used within the LHC HLT and offline computing workflows via heterogeneous computing platforms for algorithms with longer latencies.
The proof-of-principle studies discussed in Section~\ref{sec:cloud} provide an avenue toward ``Big Data'' science computing with scalability, low software maintenance cost, and maximized hardware flexibility.
%Several proof-of-principle studies deploy heterogeneous computing platforms, such as GPU and FPGA coprocessors, in HEP software with an innovative ``as-a-service'' approach~\cite{Krupa:2020bwg, Rankin:2020usv}.
%This approach provides an avenue toward ``Big Data'' science computing with scalability, low software maintenance cost, and maximized hardware flexibility.
%The SONIC software framework was developed during these studies and the SONIC-based GPU- and FPGA-as-a-service toolkits have been released as open source code.
SONIC has been integrated into the CMS experiment software stack.
The end-to-end GPU- and FPGA-based workflows for processing data in real-time heterogeneous systems is the practical approach to enable deployment of complexity algorithms needed for massive data processing.

\subsection{Neutrinos}

Machine learning algorithms are becoming increasingly prevalent and performant in the reconstruction of events in neutrino experiments. These sophisticated algorithms can be computationally expensive especially for detectors on the Earth's surface or close to a neutrino source where there are lots of activities inside the detector.
In order to improve the efficiency and speed of the inference of ML algorithms in a large-scale data processing, GPU acceleration specifically for the ProtoDUNE reconstruction chain has been integrated without disrupting the native computing workflow using SONIC~\cite{Wang:2020fjr} (as mentioned in Section~\ref{sec:cloud}). 
With the integrated framework, the most time-consuming task, track and particle shower hit identification, is accelerated by a factor of 17. This results in a factor of 2.7 reduction in the total processing time when compared with CPU-only production.
%To improve the efficiency and speed of the inference of the machine learning algorithms in a large-scale data processing, people developed a computing model in which heterogeneous computing with GPU coprocessors is made available as a web service. 
%In this approach, Services for Optimized Network Inference on Coprocessors (SONIC), GPU acceleration specifically for the ProtoDUNE reconstruction chain are integrated without disrupting the native computing workflow~\cite{Wang:2020fjr}. 
%With the integrated framework, the most time-consuming task, track and particle shower hit identification, is accelerated by a factor of 17. This results in a factor of 2.7 reduction in the total processing time when compared with CPU-only production. 
%For this particular task, only 1 GPU is required for every 68 CPU threads. 
%This initial test using a small number of simulated ProtoDUNE events showed a viable, cost-effective way to solve the computing challenge facing the neutrino experiments. In 2021, to take advantage of an improved machine learning model, the entire 7 million ProtoDUNE beam data events were reprocessed in the SONIC framework. The acceleration provided by GPUs accessed through Google Cloud reduced the total processing time by half compared to the CPU-only approach. 
Future developments are expected to include: optimizing ML algorithms for GPUs with TensorRT using approaches such as quantization; studies using other hardware such as FPGAs, IPUs (an initial test of an IPU in collaboration with GraphCore was shown to be very promising); and exploring deployment at scale at the Feynman Computing Center (at FNAL), NERSC, and other HPC centers.

%some text on trigger applications for DUNE added by Georgia

On the latency and throughput front, the future DUNE Far Detector represents a special case, where information from any contiguous region of the detector, up to several GB per region, must be processed in parallel, with millisecond latency, and selected or triggered on with high accuracy. Due to the nature of neutrino interactions, computer vision algorithms such as CNNs were the first to be explored for this task. Techniques such as network quantization have been explored to further reduce latency and resource utilization for FPGA deployment. It has been shown that a relatively simple CNN can be implemented with reasonable FPGA resource utilization to select events with sufficiently low latency and with accuracy as required by the DUNE physics program \cite{Jwa:2019zlh,Jwa:2022eaf}. Future developments include demonstration of this application using real data, either at ProtoDUNE or the Short Baseline Neutrino Detector (SBND).

\subsection{Astrophysics}

%MMA? \textbf{[1 paragraph] Erik}
%2022-03-21 Erik Katsavounidis: areas of application of ML so far and in the near future/needed
Recent work has made possible a novel implementation and deployment of a deep learning inference infrastructure for real-time gravitational-wave data analyses pipelines~\cite{Gunny:2021gne}.
This model enables easy integration of ML algorithms that can be used for denoising~\cite{Ormiston2020} and astrophysical source identification~\cite{Gabbard2018}, including the ability to scale and incorporate hardware acceleration.
Noise regression in gravitational-wave detectors is a challenging task and paramount to be performed as close to real-time as possible, since any denoising may directly improve astrophysical reach and thus discoveries that might require multi-messenger follow-up.
The ML approach to this problem enables the identification and removal of subtle features in the data, going beyond linear couplings between the gravitational-wave channel and the environment/interferometry as captured in a wealth of auxiliary witness channels.
Aside from denoising, ML algorithms for gravitational-wave source detection~\cite{Gabbard2018} and parameter estimation~\cite{Gabbard2021} are being benchmarked for full deployment during upcoming observing runs of the international network of gravitational-wave detectors.
Use of ML approaches in traditional electromagnetic astronomy has a longer history than in gravitational waves.
Supervised and unsupervised approaches have been used for object classification, including identification of artefacts, both at the pixel and light curve level.
With the rate at which data sets are growing, ML has become commonplace.
The ability to seamlessly incorporate ML models and further invoke hardware accelerators is expected to reduce the resource footprint and attain intrinsic latencies at sub-second levels.

\section{Summary and Outlook \label{sec:outlook}}

Machine learning is a powerful tool for research and offers real promise for many challenges that will face physics experiments in the coming decade.
However, ML algorithms can become very expensive computationally, and while traditional CPU-based computing will be sufficient in some cases, many others will require the use of alternative hardware to meet power, throughput, and latency constraints.
%dedicated thinking is necessary to address the present and future computing challenges that these algorithms carry.
%Many tools will be required for the wide range of needs across different frontiers and experiments.
%In some situations traditional CPU-based computing models are expected to be sufficient, while others will require the use of alternative hardware to meet power, throughput, and latency constraints.
Of the options, it seems clear that GPUs will be the choice for many needs, as they offer the most mature tools for usage and offer large speedups for most medium to large models.
FPGAs will also see use in certain situations, particularly for smaller networks and low latency applications where they are the only option such as the L1 trigger at the LHC.
Although they are not expected to play a significant role in most computing for ML in physics, the use of ASICs will be absolutely necessary in some environments like those with high radiation that simply do not allow for the use of FPGAs or GPUs.
%These architectures will all be needed to address the range of data rates and latencies across frontiers.
The field should also be prepared to adapt to new coprocessors such as neuromorphic architectures like IPUs, optical chips, or future technologies which may emerge as strong competitors to existing devices in the next decades.

In addition to the choice of hardware for ML applications, equally as important are the modes of model deployment and access.
Alternative architectures that require specialized programming languages present a significant barrier to entry for most physics users.
Compilers and transpilers from industry reduce this barrier significantly, as well as allow more efficient prototyping.
%The use of these tools and their continued development will be critical for long term success of ML in physics.
Tools from inside physics will also be critical for uses that are distinct from those targeted most by industry.
Two such tools of note are \texttt{hls4ml} and SONIC.
\texttt{hls4ml} is already in wide use for applications with ultra-low latency requirements, and its potential use cases are only expected to grow as the bounds of experiment capabilities are pushed.
SONIC and other tools to enable as-a-service computing are expected to be significant components of large computational workflows comprised of both ML and non-ML algorithms.
These tools have many advantages that make them suited not only for single physics applications but a wide range of experiments and use cases across the frontiers we have highlighted.
In all cases, both for tools from industry and from physics, it is very important that tools remain open-source to allow for collaboration and development that meets the specific needs of physics applications.

Access to the computing infrastructure necessary to execute these workflows will be critical for future success.
For specialized on- and off-detector real-time electronics, the necessity of dealing with commercial entities for licensing will necessitate communication and collaboration between laboratories and universities or the exploration of open-source solutions.
This extends to the use of hardware solutions that can be shared across experiments in the case of FPGA readout systems.
%Cloud providers and HPC centers are two major sources of alternative architectures for more generic use-cases, although both come with certain drawbacks.
As needs for experiments evolve, it is important to consider how different resources are able to scale and meet these needs.
Elastic computing offered by cloud providers is clearly scalable; as we have noted, it is currently costly for sustained usage, but costs may decrease in the future.
Collaboration with HPC centers will be important to ensure they can be utilized in ways most effective for experimental workflows that may not require their baseline of 2:1 ratios of CPU to accelerator.

The next decade will present significant challenges to physics research, and ML will be a necessary component in overcoming some of these challenges.
The tools and resources necessary to enable these ML solutions will come both from physics and industry, and will come in the form of both dedicated hardware and software and computing paradigms.
Continued collaboration with industry and HPC centers will be critical to handling the rapidly changing landscape of ML.
This can enable the not only the exciting applications discussed here but also those that are to come in the future.

\bibliographystyle{JHEP}
\bibliography{references}

\end{document}